\documentclass[a4paper]{article}
\usepackage[top=4cm, bottom=4cm, left=2cm, right=2cm]{geometry} 

\usepackage{amssymb}
\usepackage{amsthm}

\usepackage{amsmath}

\usepackage{float}
\usepackage{graphicx}
\usepackage{url}

\newcommand{\comment}[1]{}

\newfloat{algorithm}{tbp}{loa}
\floatname{algorithm}{Algorithm}

\usepackage{fancyhdr}
\usepackage{xspace}
\usepackage{times}
\newcommand{\eprime}{{\sc Essence}$'$\xspace}
\newcommand{\essence}{{\sc Essence}\xspace}
\newcommand{\savilerow}{{\sc Savile Row}\xspace}

\newcommand{\version}{1.6.4\xspace}

\begin{document}

\title{\eprime Description \version}

\author{Peter Nightingale and Andrea Rendl}

\date{}

\maketitle

\section{Introduction}

The purpose of this document is to describe the \eprime language and to be a 
reference for users of \eprime.  \eprime is a constraint modelling language,
therefore it is mainly designed for describing $\mathcal{NP}$-hard decision 
problems. It is not the only (or the first) constraint modelling language. 
\eprime began as a subset of \essence~\cite{essence-journal-08} and has been extended
from there. It is similar to the earlier Optimization Programming Language (OPL)~\cite{opl-book}.
\eprime is implemented by the tool \savilerow~\cite{nightingale2014automatically, nightingale2015automatically}.

\eprime is considerably different to procedural programming languages, it does not specify a procedure
to solve the problem. The user specifies the problem in terms of decision variables
and constraints, and the solver automatically finds a value for each variable 
such that all constraints are satisfied. This means, for example, that the 
order the constraints are presented in \eprime is irrelevant. 

\begin{figure}
\begin{center}
\includegraphics[width=0.3\textwidth]{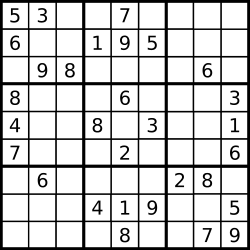}\:\:\includegraphics[width=0.3\textwidth]{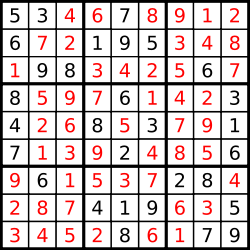}
\end{center}
\caption{\label{fig:sudoku}On the left is a Sudoku puzzle. The objective is to fill in all blank
squares using only the digits 1-9 such that each row contains all the digits 1-9, each column 
also contains all the digits 1-9, and each of the 3 by 3 subsquares outlined in thicker lines also 
contains all digits 1-9. 
On the right is the solution. 
(Images are public domain from Wikipedia.)
}
\end{figure}

\eprime allows the user to solve \textit{constraint satisfaction problems} (CSPs). 
A simple example of a CSP is a Sudoku puzzle (Figure~\ref{fig:sudoku}). To convert
a single Sudoku puzzle to CSP, each square can be represented as a decision
variable with domain $\{1\ldots 9\}$. The clues (filled in squares) and the rules of the puzzle are
easily expressed as constraints. 

We will use Sudoku as a running example. A simple first attempt of modelling Sudoku
in \eprime is shown below. In this case the clues (for example \texttt{M[1,1]=5}) 
are included in the model. We have used not-equal constraints to state that all 
digits must be used in each row and column. We have also omitted the sub-square
constraints for now. 

\begin{verbatim}
language ESSENCE' 1.0

find M : matrix indexed by [int(1..9), int(1..9)] of int(1..9)

such that

M[1,1]=5,
M[1,2]=3,
M[1,5]=7,
....
M[9,9]=9,

forAll row : int(1..9) . 
    forAll col1 : int(1..9) .
        forAll col2: int(col1+1..9) . M[row, col1]!=M[row, col2],

forAll col : int(1..9) . 
    forAll row1 : int(1..9) .
        forAll row2: int(row1+1..9) . M[row1, col]!=M[row2, col]

\end{verbatim}

In this example, some CSP decision variables are declared using a \texttt{find} statement. 
It is also worth noting that other variables exist that are not decision variables,
for example, \texttt{row} is a \textit{quantifier} variable that exists only to apply some 
constraints to every row. 

An \eprime model usually describes a \textit{class} of CSPs. For example, it 
might describe the class of all Sudoku puzzles. In order to solve a particular
instance of Sudoku, the instance would be specified in a separate \textit{parameter} file (also 
written in \eprime). The model would have parameter variables (of type integer, boolean or matrix), 
and the parameter file would specify a value for each of these variables.

Since \eprime is a constraint modelling language, we will define the
constraint satisfaction problem (CSP). 
A CSP $\mathcal{P}=\langle\mathcal{X},\mathcal{D},\mathcal{C}\rangle$
is defined as a set of $n$ \textit{decision variables} 
$\mathcal{X}=\langle x_{1},\dots,x_{n}\rangle$,
a set of domains $\mathcal{D}=\langle D(x_{1}),\dots,D(x_{n})\rangle$ where
$D(x_{i})\subsetneq\mathbb{Z}$, $\left|D(x_{i})\right|<\infty$ is the
finite set of all potential values of $x_{i}$, and a conjunction
$\mathcal{C}=C_{1}\wedge C_{2}\wedge\cdots\wedge C_{e}$ of constraints. 

For CSP $\mathcal{P}=\langle\mathcal{X},\mathcal{D},\mathcal{C}\rangle$,
a constraint $C_{k}\in\mathcal{C}$ consists of a sequence of $m>0$
variables $\mathcal{X}_{k}=\langle x_{k_{1}},\dots,x_{k_{m}}\rangle$
with domains $\mathcal{D}_{k}=\langle D(x_{k_{1}}),\ldots,D(x_{k_{m}})\rangle$
s.t. $\mathcal{X}_{k}$ is a subsequence%
\footnote{I use subsequence in the sense that $\langle1,3\rangle$ is a subsequence 
of $\langle1,2,3,4\rangle$.%
} of $\mathcal{X}$, $\mathcal{D}_{k}$ is a subsequence of $\mathcal{D}$,
and each variable $x_{k_{i}}$ and domain $D(x_{k_{i}})$ matches a variable
$x_{j}$ and domain $D(x_{j})$ in $\mathcal{P}$. $C_{k}$ has an associated
set $C_{k}^{S}\subseteq D(x_{k_{1}})\times\dots\times D(x_{k_{m}})$ of
tuples which specify allowed combinations of values for the variables
in $\mathcal{X}_{k}$.

\comment{
forAll subsqrow : int(1,4,7) .      $ subsqrow and subsqcol are the top-left corner 
    forAll subsqcol : int(1,4,7) .  $ of the subsquare. 
        forAll row1 : int(1..3) . 
            forAll row2 : int(1..3) .
                forAll col1 : int(1..3) .
                    forAll col2 : int(1..3) .
                        (row1!=row2 \/ col1!=col2) -> M[subsqrow+row1,subsqcol+col1]!=M[subsqrow+row2,subsqcol+col2]
}

\section{The \eprime Expression Language}

In \eprime \textit{expressions} are built up from variables and literals using 
operators (such as \texttt{+}). We will start by describing the types
that expressions may take, then describe the simplest expressions and
build up from there. 

\subsection{Types and Domains}\label{sec:types-domains}

Types and domains play a similar role; they prescribe a set of
values that an expression or variable can take. Types denote non-empty sets that contain all elements
that have a similar structure, whereas domains denote possibly empty
sets drawn from a single type.  In this manner, each domain is associated with an
underlying type.  For example integer is the type underlying the
domain comprising integers between 1 and 10. The type contains all integers, 
and the domain is a finite subset. 

\eprime is a strongly typed language;
every expression has a type, and the types of all
expressions can be inferred and checked for correctness.
Furthermore, \eprime is a finite-domain language; every decision variable is
associated with a finite domain of values. 

The atomic types of \eprime are {\tt int} (integer) and {\tt bool} (boolean). 
There is also a compound type, \texttt{matrix},
that is constructed of atomic types.

There are three different types of domains in \eprime{}:
boolean, integer and matrix domains. Boolean and integer 
domains are both atomic domains; matrix domains are built from 
atomic domains.

\begin{description}

\item[Boolean Domains]
{\tt bool} is the Boolean domain consisting of {\tt false} and {\tt true}.

\item[Integer Domains]
An integer domain represents a finite subset of the integers, and is specified 
as a sequence of ranges or individual elements, for example 
{\tt int(1..3,5,7,9..20)}. Each range includes the endpoints, so the meaning of a range {\tt a..b} is the set 
\(\{ i \in \mathbb{Z} | a\le i \le b \}\). A range {\tt a..b} would normally be 
in order, i.e.~{\tt a<=b} but this is not strictly required. Out-of-order ranges 
correspond to the empty set of integers. 

The meaning of an integer domain is the union of the ranges and individual elements in the domain. 
For example, {\tt int(10, 1..5, 4..9)} is equivalent to {\tt int(1..10)}. 

Finally, the elements and endpoints of ranges may be expressions of type {\tt int}. 
The only restriction is that they may not contain any CSP decision variables.  
The integer expression language is described in the following sections.

\item[Matrix Domains]
A matrix is defined by the keyword {\tt matrix}, followed by its dimensions and 
the base domain. The dimensions are a list, in square brackets, of domains.
For instance,
\begin{verbatim}
Matrix1 : matrix indexed by [int(1..10),int(1..10)] of int(1..5) 
\end{verbatim}
is the domain of a two-dimensional square matrix, indexed by $1..10$ in both dimensions,
where each element of the matrix has the domain {\tt int(1..5)}. Elements of this matrix would 
be accessed by {\tt Matrix1[A,B]} where A and B are integer expressions. 

Matrices may be indexed by any integer domain or the boolean domain. For example, 

\begin{verbatim}
Matrix2 : matrix indexed by [int(-2..5),int(-10..10,15,17)] of int(1..5)
\end{verbatim}

is a valid matrix domain.

\end{description}

\subsection{Domain Expressions}\label{sec:domainexpressions}

\eprime contains a small expression language for boolean and integer domains. 
This language consists of three binary infix operators \texttt{intersect}, \texttt{union} and \texttt{-}.
All three are left-associative and the precedences are given in Appendix~\ref{app:op}.
The language also allows bracketed subexpressions with \texttt{(} and \texttt{)}. 

For example, the first and second lines below are exactly equivalent. 

\begin{verbatim}
letting dom be domain int(1..5, 3..8)
letting dom be domain int(1..5) union int(3..8)
\end{verbatim}

\subsection{Literals}\label{sec:literals}

Each of the three types (integer, boolean and matrix) has a corresponding literal 
syntax in \eprime. Any value of any type may be written as a literal. Sets and 
real numbers are not (as yet) part of the language.  Integer and boolean literals
are straightforward:

\begin{center}
\begin{verbatim}
1 2 3 -5
true false
\end{verbatim}
\end{center}

There are two forms of matrix literals. The simpler form is a comma-separated list of
expressions surrounded by square brackets. For example, the following is a matrix
literal containing four integer literals. 

\begin{center}
{\tt [ 1, 3, 2, 4 ]}
\end{center}

Matrix literals may contain any valid expression in \eprime. For example a
matrix literal is allowed to contain other matrix literals to build up a matrix 
with two or more dimensions. The types of the expressions contained in the matrix
literal must all be the same. 

The second form of matrix literal has an explicit index domain that specifies how 
the literal is indexed. This is specified after the comma-separated list of
contents using a \texttt{;} as follows.

\begin{center}
{\tt [ 1, 3, 2, 4 ; int(4..6,8) ]}
\end{center}

The index domain must be a domain of type \texttt{bool} or \texttt{int}, and must 
contain the same number of values as the number of items in the matrix literal. 
When no index domain is specified, a matrix of size \texttt{n} is indexed by \texttt{int(1..n)}. 

Finally a multi-dimensional matrix value can be expressed by nesting matrix literals.
Suppose we have the following domain:

\begin{center}
\texttt{matrix indexed by [int(-2..0),int(1,2,4)] of int(1..5)}
\end{center}

One value contained in this domain is the following:

\begin{verbatim}
[ [ 1,2,3 ; int(1,2,4) ],
  [ 1,3,2 ; int(1,2,4) ],
  [ 3,2,1 ; int(1,2,4) ]
  ; int(-2..0) ]
\end{verbatim}

\subsection{Variables}\label{sec:vars}

Variables are identified by a string. Variable names must start with a letter 
\texttt{a-z} or \texttt{A-Z}, and after the first letter may contain any of
the following characters: \texttt{a-z A-Z 0-9 \_}. A variable may be of type
integer, boolean or matrix. 

Scoping of variables depends on how they are declared and is dealt with in the
relevant sections below. As well as a type, variables have a category. The 
category is \textit{decision}, \textit{quantifier} or \textit{parameter}. Decision
variables are CSP variables, and the other categories are described below. 

Expressions containing decision variables are referred to as \textit{decision expressions},
and expressions containing no decision variables as \textit{non-decision expressions}.
This distinction is important because expressions in certain contexts are not allowed
to contain decision variables. 

\subsection{Expression Types}

Expressions can be of any of the three basic types (integer, boolean or matrix). 
Integer expressions range over an integer domain, 
for instance {\tt x + 3} (where \texttt{x} is an integer variable) is an integer expression ranging from 
$lb(x)+3$ to $ub(x)+3$. Boolean expressions range over the 
Boolean domain, for instance the integer comparison 
 {\tt x = 3} can either be {\tt true}
or {\tt false}. 

\subsection{Type Conversion}

Boolean expressions or literals are automatically converted to integers when used
in a context that expects an integer. As is conventional \texttt{false} is converted
to \texttt{0} and \texttt{true} is converted to {1}. For example, the following are 
valid \eprime boolean expressions. 

\begin{verbatim}
1+2-3+true-(x<y)=5
false < true
\end{verbatim}

Integer expressions are not automatically converted to boolean. Matrix expressions
cannot be converted to any other type (or vice versa). 

\subsection{Matrix Indexing and Slicing}\label{sub:matrix-slicing}

Suppose we have a three-dimensional matrix {\tt M} with the following domain:

\begin{verbatim}
matrix indexed by [int(1..3), int(1..3), bool] of int(1..5)
\end{verbatim}

{\tt M} would be indexed as follows: {\tt M[x,y,z]}, where \texttt{x} and \texttt{y} may be integer or
boolean expressions and \texttt{z} must be boolean. Because the matrix has the base
domain \texttt{int(1..5)}, {\tt M[x,y,z]} will be an integer expression. 
Matrix indexing is a partial function: when one of the indices is out of bounds 
then the expression is undefined. \eprime has the relational semantics, 
in brief this means that the boolean expression containing the undefined expression is
\texttt{false}. So for example, \texttt{M[1,1,false]=M[2,4,true]} is always \texttt{false}
because the 4 is out of bounds. The relational semantics are more fully described 
in Section~\ref{sec:undef} below. 

Parts of matrices can be extracted by \textit{slicing}. Suppose we have the 
following two-dimensional matrix named \texttt{N}: 

\begin{verbatim}
[ [ 1,2,3 ; int(1,2,4) ],
  [ 1,3,2 ; int(1,2,4) ],
  [ 3,2,1 ; int(1,2,4) ]
  ; int(-2..0) ]
\end{verbatim}

We could obtain the first row by writing \texttt{N[-2,..]}, which is equal to \texttt{[ 1,2,3 ; int(1..3)]}. 
Similarly the first column can be obtained by 
\texttt{N[..,1]} which is \texttt{[ 1,1,3 ; int(1..3)]}.  In general, the indices in a matrix slice may be \texttt{..} or 
an integer or boolean expression that does not contain any decision variables. 
Matrix slices are always indexed contiguously from 1 regardless of the original matrix
domain. 

When a matrix slice has an integer or boolean index that is out of bounds, then
the expression is undefined and this is handled as described in Section~\ref{sec:undef}.

\subsection{Integer and Boolean Expressions}\label{sec:int-bool-expressions}

\eprime has a range of binary infix and unary operators and functions for building up 
integer and boolean expressions, for example:

\begin{itemize}
\item Integer operators: {\tt + - * ** / \% | min max}
\item Boolean operators: {\tt \verb1\/ /\1 -> <-> !}
\item Quantified sum: {\tt sum}
\item Logical quantifiers: {\tt forAll exists}
\item Numerical comparison operators: {\tt = != > < >= <=}
\item Matrix comparison operators: {\tt <=lex <lex >=lex >lex}
\item Set operator: {\tt in}
\item Global constraints: {\tt allDiff gcc}
\item Table constraint: {\tt table}
\end{itemize}

These are described in the following subsections. 

\subsubsection{Integer Operators}

\eprime has the following binary integer 
operators: {\tt + - * / \% **}. {\tt +}, {\tt -} and {\tt *} are the standard
integer operators. 

The operators \texttt{/} and \texttt{\%} are integer division and modulo functions. 
{\tt a/b} is defined as $\lfloor a/b \rfloor$ (i.e. it always rounds down). 
This does not match some programming languages, for example C99 which rounds 
towards 0. 

The modulo operator {\tt a\%b} is defined as $a-b\lfloor a/b \rfloor$ to be 
complementary to \texttt{/}.

\begin{verbatim}
3/2 = 1
(-3)/2 = -2
3/(-2) = -2
(-3)/(-2) = 1

3 % 2 = 1
(-3) % 2 = 1
3 % (-2) = -1
(-3) % (-2) = -1
\end{verbatim}

{\tt **} is the power function: {\tt x**y} is defined as $x^y$.
There are two unary functions: absolute value (where {\tt |x|} is the absolute 
value of {\tt x}), and negation (unary {\tt -}).

\subsubsection{Boolean Operators}

\eprime has the {\tt \verb1/\1} (and) and {\tt \verb1\/1} (or) operators 
defined on boolean expressions. There are also {\tt ->} (implies), {\tt <->} 
(if and only if), and {\tt !} (negation). These operators all take boolean 
expressions and produce a new boolean expression. They can be nested arbitrarily.  

The comma {\tt ,} in \eprime is also a binary boolean operator, with the same meaning 
as {\tt \verb1/\1}. However it has a different precedence, and is used quite 
differently. {\tt \verb1/\1} is normally used within a constraint, and
{\tt ,} is used to separate constraints (or separate groups of constraints 
constructed using a \texttt{forAll}). Consider the following example.

\begin{verbatim}
forAll i : int(1..n) . x[i]=y[i] /\ x[i]!=y[i+1],
exists i : int(1..n) . x[i]=1 /\ y[i]!=y[i+1]
\end{verbatim}

Here we have two quantifiers, both with an {\tt \verb1/\1} inside. The comma is 
used to separate the {\tt forAll} and the {\tt exists}. The comma has the lowest precedence of all binary operators.

\subsubsection{Integer and Boolean Functions}

\eprime has named functions {\tt min(X,Y)} and {\tt max(X,Y)} that both take 
two integer expressions \texttt{X} and \texttt{Y}. 
\texttt{min} and \texttt{max} can also be applied to one-dimensional matrices to
obtain the minimum or maximum of all elements in the matrix (see Section~\ref{sec:matrix-functions}).
\texttt{factorial(x)} returns the factorial of values from 0 to 20 where the result fits in
a 64-bit signed integer. It is undefined for other values of \texttt{x} and the expression \texttt{x} is not 
allowed to contain decision variables. \texttt{popcount(x)} returns the bit count of the 
64-bit two's complement representation of \texttt{x}, and \texttt{x} may not contain 
decision variables. 

The function \texttt{toInt(x)} takes a boolean expression \texttt{x} and 
converts to an integer 0 or 1. This function is included only for compatibility with 
\textsc{Essence}: it is not needed in \eprime because booleans are automatically cast to
integers. 

\subsubsection{Numerical Comparison Operators}

\eprime provides the following integer comparisons with their obvious meanings: 
{\tt = != > < >= <=}. These operators each take two integer expressions and produce
a boolean expression. 

\subsubsection{Matrix Comparison Operators}

\eprime provides a way of comparing one-dimensional matrices. These operators compare two 
matrices using the dictionary order (\textit{lexicographical} order, or lex for 
short). 

There are four operators. {\tt A <lex B} ensures that A comes before B in 
lex order, and {\tt A <=lex B} which ensures that {\tt A <lex B} or {\tt A=B}.
{\tt >=lex} and {\tt >lex} are identical but with the arguments reversed. 
For all four operators, A and B may have different lengths and may be indexed differently, but they must be
one-dimensional.  Multi-dimensional matrices may be flattened to one dimension using
the \texttt{flatten} function described in Section~\ref{sec:matrix-functions} below. 

\subsubsection{Set Operator}

The operator \texttt{in} states that an integer expression takes a value in a 
set expression. The set espression may not contain decision variables. 
The set may be a domain expression 
(Section~\ref{sec:domainexpressions}) or the \texttt{toSet} function that
converts a matrix to a set, as in the examples below. 

\begin{verbatim}
x+y in (int(1,3,5) intersect int(3..10))
x+y in toSet([ i | i : int(1..n), i%2=0])
\end{verbatim}

\subsubsection{The Quantified Sum Operator}\label{sub:qsum}

The {\tt sum} operator corresponds to the mathematical 
$\sum$ and has the following syntax:
\begin{center}
{\tt sum i : D . E} 
\end{center}
where {\tt i} is a quantifier variable, {\tt D} is a domain, and {\tt E} is the
expression contained within the {\tt sum}. More than one quantifier variable may
be created by writing a comma-separated list {\tt i,j,k}. 

For example, if we want to take the sum of all numbers in the range 1 to 10 we write

\begin{verbatim}
sum i : int(1..n) . i
\end{verbatim}

which corresponds to $\sum\nolimits_{i=1}^n i$. {\tt n} cannot be a decision 
variable.

Quantified sum has several similarities to the {\tt forAll} and {\tt exists} 
quantifiers (described below in Section~\ref{sub:forallexists}): it introduces
new local variables (named quantifier variables) that can be used within {\tt E}, 
and the quantifier variables all have the same domain {\tt D}. However {\tt sum} 
has one important difference: a {\tt sum} is an integer expression. 

A quantified sum may be nested inside any other integer operator, including another
quantified sum:

\begin{verbatim}
sum i,j : int(1..10) .
    sum k : int(i..10) .
        x[i,j] * k
\end{verbatim}

\subsubsection{Universal and Existential Quantification} \label{sub:forallexists}

Universal and existential quantification are powerful means to 
write down a series of constraints in a compact way. 
Quantifications have the same syntax as {\tt sum}, but with 
{\tt forAll} and {\tt exists} as keywords:

\begin{verbatim}
forAll  i : D . E
exists  i : D . E
\end{verbatim}

For instance, the universal quantification

\begin{verbatim}
forAll i : int(1..3) . x[i] = i
\end{verbatim}

corresponds to the conjunction:

\begin{verbatim}
x[1] = 1 /\ x[2] = 2 /\ x[3] = 3
\end{verbatim}

An example of existential quantification is

\begin{verbatim}
exists i : int(1..3) . x[i] = i
\end{verbatim}

and it corresponds to the following disjunction:

\begin{verbatim}
x[1] = 1 \/ x[2] = 2 \/ x[3] = 3
\end{verbatim}

Quantifications can range over several quantified variables and can 
be arbitrarily nested, as demonstrated with the {\tt sum} quantifier.

In the running Sudoku example, \texttt{forAll} quantification is used to build
the set of constraints. The expression:

\begin{verbatim}
forAll row : int(1..9) . 
    forAll col1 : int(1..9) .
        forAll col2: int(col1+1..9) . M[row, col1]!=M[row, col2]
\end{verbatim}

is a typical use of universal quantification. 

\subsubsection{Quantification over Matrix Domains}

All three quantifiers are defined on matrix domains as well as integer and 
boolean domains. For example, to quantify over all permutations of size \texttt{n}:

\begin{verbatim}
forAll perm : matrix indexed by [int(1..n)] of int(1..n) . 
    allDiff(perm) -> exp
\end{verbatim}

The variable \texttt{perm} represents a matrix drawn from the matrix domain, and the 
\texttt{allDiff} constraint evaluates to \texttt{true} when \texttt{perm} is a 
permutation of $1\ldots n$. Hence the expression \texttt{exp} is quantified for
all permutations of $1\ldots n$. 

If \texttt{n} is a constant, the example above could be written as a set of \texttt{n}
nested \texttt{forAll} quantifiers. However if \texttt{n} is a parameter of the 
problem class, it is very difficult to write the example above using other (non-matrix) 
quantifiers. 

\subsubsection{Global Constraints}

\eprime provides a small set of global constraints such as {\tt allDiff} (which is satisfied when a
vector of variables each take different values). Global constraints are all
boolean expressions in \eprime. 
Typically it is worth using these in models because the solver often performs
better with a global constraint compared to a set of simpler constraints. 

For example, the following two lines are semantically equivalent (assuming
{\tt x} is a matrix indexed by {\tt 1..n}). 

\begin{verbatim}
forAll i,j : int(1..n) . i<j -> x[i]!=x[j]
allDiff(x)
\end{verbatim}

Both lines will ensure that the variables in {\tt x} take different values. 
However the {\tt allDiff} will perform better in most situations.\footnote{In the 
current version of Savile Row, with default settings, the \texttt{x[i]!=x[j]} constraints would be aggregated to create \texttt{allDiff(x)} therefore there is no difference in performance between these two statements. 
There would be a difference in performance when aggregation is switched off (for example by using the \texttt{-O1} flag).}
Table \ref{tab:globals} summarises the global constraints available in \eprime.

\begin{table}  
    \begin{center}
    \begin{tabular}{l|l|l} 
Global Constraint & Arguments & Description \\
\hline
{\tt allDiff(X)} & {\tt X} is a matrix & Ensures expressions in {\tt X} take \\
                      &                     & distinct values in any solution.  \\
\hline
{\tt atleast(X, C, Vals)} & {\tt X} is a matrix & For each non-decision expression \\
                        & {\tt Vals} is a matrix of non-decision expressions & {\tt Vals[i]}, the number of occurrences  \\
                      & {\tt C} is a matrix of non-decision expressions & of {\tt Vals[i]} in {\tt X} is at least {\tt C[i]}.   \\
\hline
{\tt atmost(X, C, Vals)} & {\tt X} is a matrix & For each non-decision expression \\
                        & {\tt Vals} is a matrix of non-decision expressions & {\tt Vals[i]}, the number of occurrences  \\
                      & {\tt C} is a matrix of non-decision expressions & of {\tt Vals[i]} in {\tt X} is at most {\tt C[i]}.   \\
\hline
{\tt gcc(X, Vals, C)} & {\tt X} is a matrix & For each non-decision expression \\
                      & {\tt Vals} is a matrix of non-decision expressions & {\tt Vals[i]}, the number of occurrences  \\
                      & {\tt C} is a matrix & of {\tt Vals[i]} in {\tt X} equals {\tt C[i]}.   \\
\hline
{\tt alldifferent\verb1_1except} & {\tt X} is a matrix  & Ensures variables in {\tt X} take \\
\quad {\tt (X, Value)}           & {\tt Value} is a non-decision expression & distinct values, except that \\
                               &                           & {\tt Value} may occur any number \\
                               &                           & of times.\\ 
      \end{tabular}
\end{center}
\caption{Global constraints in \eprime. Each may be nested within expressions and have arbitrary expressions
    nested within them. Each matrix parameter of a global constraint must be a one-dimensional
    matrix. In some cases the parameter is a matrix of \textit{non-decision expressions} -- that is, integer or 
    boolean expressions that contain no decision variables. Quantifier and parameter variables are allowed
    in non-decision expressions. }
\label{tab:globals}
\end{table}

Now we have the \texttt{allDiff} global constraint, the Sudoku example can be improved
and simplified as follows:

\begin{verbatim}
language ESSENCE' 1.0

find M : matrix indexed by [int(1..9), int(1..9)] of int(1..9)

such that

M[1,1]=5,
M[1,2]=3,
M[1,5]=7,
....
M[9,9]=9,

forAll row : int(1..9) .
    allDiff(M[row,..]),

forAll col : int(1..9) .
    allDiff(M[..,col])	 
\end{verbatim}

Global constraints are boolean expressions like any other, and are allowed to be
used in any context that accepts a boolean expression. 

\subsubsection{Table Constraints}

In a table constraint the satisfying tuples of the constraint are specified using a matrix. 
This allows a table constraint to theoretically implement any relation, although
practically it is limited to relations where the set of satisfying tuples is small
enough to store in memory and efficiently search. 

The first argument specifies the variables in the scope of the
constraint, and the second argument is a two-dimensional matrix of satisfying tuples. 
For example, the constraint {\tt a+b=c} on boolean variables could be written
as a table as follows. 

\begin{verbatim}
table( [a,b,c], [[0,0,0], [0,1,1], [1,0,1]] )
\end{verbatim}

The first argument of \texttt{table} is a one-dimensional matrix expression. It
may contain both decision variables and constants. The second argument is a 
two-dimensional matrix expression containing no decision variables. The second argument
can be stated as a matrix literal, or an identifier, or by slicing a higher-dimension
matrix, or by constructing a matrix using matrix comprehensions (see Section~\ref{sec:comprehensions}). 

If the same matrix of tuples is used for many table constraints, a {\tt letting} 
statement can be used to state the matrix once and use it many times. Lettings are
described in Section~\ref{sec:letting} below. 

\subsection{Matrix Comprehensions}\label{sec:comprehensions}

We have seen that matrices may be written explicitly as a matrix literal (Section~\ref{sec:literals}), and 
that existing matrices can be indexed and sliced (Section~\ref{sub:matrix-slicing}).  Matrices
can also be constructed using \textit{matrix comprehensions}. This provides a 
very flexible way to create matrices of variables or values. A single 
matrix comprehension creates a one-dimensional matrix, however they can be
nested to create multi-dimensional matrices. There are two syntactic forms of matrix
comprehension:

\begin{verbatim}
[ exp | i : domain1, j : domain2, cond1, cond2 ]
[ exp | i : domain1, j : domain2, cond1, cond2 ; indexdomain ]
\end{verbatim}

where \texttt{exp} is any integer, boolean or matrix expression. 
This is followed by any number of comprehension variables, each with a domain.
After the comprehension variables we have an optional list of conditions: these
are boolean expressions that constrain the values of the comprehension variables. 
Finally there is an optional index domain. This provides an index domain to the 
constructed matrix. 

To expand the comprehension, each assignment to the comprehension 
variables that satisfies the conditions is enumerated in lexicographic order. 
For each such assignment, the values of the comprehension variables are substituted into \texttt{exp}.
The resulting expression then becomes one element of the constructed matrix.

The simplest matrix comprehensions have only one comprehension variable, as in the example
below. 

\begin{verbatim}
[ num**2 | num : int(1..5) ] = [ 1,4,9,16,25 ; int(1..5) ]
\end{verbatim}

The matrix constructed by a comprehension is one-dimensional and is either indexed
from 1 contiguously, or has the given index domain. The given domain must have a lower bound
but is allowed to have no upper bound. 
For example the first line below produces the matrix on the second line. 

\begin{verbatim}
[ i+j | i: int(1..3), j : int(1..3), i<j ; int(7..) ]
[ 3, 4, 5 ; int(7..9) ]
\end{verbatim}

Matrix comprehensions allow more advanced forms of slicing than the matrix slice syntax in
Section~\ref{sub:matrix-slicing}. For example it is possible to slice an arbitrary subset of the rows or 
columns of a two-dimensional matrix. The following two nested comprehensions will 
slice out the entries of a matrix \texttt{M} where both rows and columns are odd-numbered, 
and build a new two-dimensional matrix. 

\begin{verbatim}
[ [ M[i,j] | j : int(1..n), j%2=1 ] | i : int(1..n), i%2=1 ]
\end{verbatim}

Now we have matrix comprehensions, the Sudoku example can be completed by adding
the constraints on the $3\times 3$ subsquares. A comprehension is used to 
create a matrix of variables and the matrix is used as the parameter of an
\texttt{allDiff} constraint. 

\begin{verbatim}
language ESSENCE' 1.0

find M : matrix indexed by [int(1..9), int(1..9)] of int(1..9)

such that

M[1,1]=5,
M[1,2]=3,
M[1,5]=7,
....
M[9,9]=9,

forAll row : int(1..9) .
    allDiff(M[row,..]),

forAll col : int(1..9) .
    allDiff(M[..,col]),	 

$ all 3x3 subsquare have to be all-different
$ i,j indicate the top-left corner of the subsquare. 
forAll i,j : int(1,4,7) .
    allDiff([ M[k,l]  | k : int(i..i+2), l : int(j..j+2)])
\end{verbatim}

In this example, the matrix constructed by the comprehension depends on the values 
of \texttt{i} and \texttt{j} from the \texttt{forAll} quantifier.  The comprehension 
variables \texttt{k} and \texttt{l} each take one of three values, to cover the 9 entries \texttt{M[k,l]} in the 
subsquare. 

\subsubsection{Matrix Comprehensions over Matrix Domains}

Similarly to quantifiers, matrix comprehension variables can have a matrix domain.
For example, the following comprehension builds a two-dimensional matrix where
the rows are all permutations of \texttt{1..n}. 

\begin{verbatim}
[ perm | perm : matrix indexed by [int(1..n)] of int(1..n), allDiff(perm) ]
\end{verbatim}

\subsection{Functions on Matrices\label{sec:matrix-functions}}

Table \ref{tab:matfunc} lists the matrix functions available in \eprime. 

\begin{table}
 \begin{tabular}{l|l|l} 
Function & Arguments & Description \\
\hline
{\tt sum(X)} & {\tt X} is a one-dimensional matrix & Constructs the sum of elements in \texttt{X} \\
\hline
{\tt product(X)} & {\tt X} is a one-dimensional matrix & Constructs the product of elements in \texttt{X} \\
\hline
{\tt and(X)} & {\tt X} is a one-dimensional matrix of booleans & Constructs the conjunction of \texttt{X} \\
\hline
{\tt or(X)} & {\tt X} is a one-dimensional matrix of booleans & Constructs the disjunction of \texttt{X} \\
\hline
{\tt min(X)} & {\tt X} is a one-dimensional matrix & The integer minimum of elements in \texttt{X} \\
\hline
{\tt max(X)} & {\tt X} is a one-dimensional matrix & The integer maximum of elements in \texttt{X} \\
\hline
{\tt flatten(X)} & {\tt X} is a matrix & Constructs a one-dimensional matrix (indexed  \\
                 &                     & contiguously from 1) with the same contents as \texttt{X} \\
\hline
{\tt flatten(n,X)} & {\tt X} is a matrix & The first \texttt{n+1} dimensions of \texttt{X} are \\
                   &                     & flattened into one dimension that is indexed \\
                   &                     & contiguously from 1. Therefore \texttt{flatten(n,X)} \\
                   &                     & produces a new matrix with \texttt{n} fewer dimensions\\
                   &                     & than \texttt{X}. The first argument \texttt{n} must be positive\\
                   &                     & or 0, and \texttt{flatten(0,X)} returns \texttt{X} unchanged.\\
\hline
{\tt toSet(X)} & {\tt X} is a one-dimensional matrix & The set of elements in \texttt{X} \\
               & of non-decision expressions         & \\
\end{tabular}
\caption{Matrix Functions\label{tab:matfunc}}
\end{table}

The functions \texttt{sum}, \texttt{product}, \texttt{and} and \texttt{or} were originally
intended to be used with matrix comprehensions, but can be used with any matrix. 
The quantifiers \texttt{sum}, \texttt{forAll} and \texttt{exists}
can be replaced with \texttt{sum}, \texttt{and} and \texttt{or} containing matrix comprehensions. 
For example, consider the \texttt{forAll} expression below (from the Sudoku model). It can be replaced with
the second line below. 

\begin{verbatim}
forAll row : int(1..9) . allDiff(M[row,..])

and([ allDiff(M[row,..]) | row : int(1..9) ])
\end{verbatim}

In fact, matrix functions combined with matrix comprehensions are strictly more
powerful than quantifiers. Also, the function \texttt{product} has no corresponding
quantifier. Quantifiers are retained in the language because they can be easier to read.

As a more advanced example, given an $n\times n$ matrix \texttt{M} of decision variables, the sum below
is the determinant of \texttt{M} using the Leibniz formula. The outermost comprehension
constructs all permutations of $1\ldots n$ using a matrix domain and an \texttt{allDiff}.
Lines 3 and 4 contain a comprehension that is used to obtain the number of \textit{inversions}
of \texttt{perm} (the number of pairs of values that are not in ascending order). 
Finally line 5 builds a product of some of the entries of the matrix. 
Without the \texttt{product} function, it is difficult (perhaps impossible) to write the Leibniz formula
in \eprime for a matrix of unknown size $n$. 

\begin{verbatim}
sum([
    $  calculate the sign of perm from the number of inversions. 
    ( (-1) ** sum([ perm[idx1]>perm[idx2] 
                  | idx1 : int(1..n), idx2 : int(1..n), idx1<idx2 ]) )*   
    product([ M[j, perm[j]] | j : int(1..n) ])
| perm : matrix indexed by [int(1..n)] of int(1..n), allDiff(perm)])
\end{verbatim}

The \texttt{flatten} function is typically used to feed the contents of a 
multi-dimensional matrix expression into a constraint that accepts only 
one-dimensional matrices. For example, given a three-dimensional matrix \texttt{M},
the following example is a typical use of \texttt{flatten}.  

\begin{verbatim}
allDiff( flatten(M[1,..,..]) )
\end{verbatim}

When flattening a matrix \texttt{M} to create a new matrix \texttt{F}, the order
of elements in \texttt{F} is defined as follows. Suppose \texttt{M} were written as
a matrix literal (as in Section~\ref{sec:literals}) the order elements are written
in the matrix literal is the order the elements appear in \texttt{F}. The following
example illustrates this for a three-dimensional matrix. 

\begin{verbatim}
flatten([ [ [1,2], [3,4] ], [ [5,6], [7,8] ] ]) = [1,2,3,4,5,6,7,8]
\end{verbatim}

\section{Model Structure}

An \eprime model is structured in the following way:

\begin{enumerate}
\item Header with version number: {\tt  language ESSENCE' 1.0} 
\item Parameter declarations (optional)
\item Constant definitions (optional)
\item Decision variable declarations (optional)
\item Constraints on parameter values (optional)
\item Objective (optional)
\item Solver Control (optional)
\item Constraints
\end{enumerate}

Parameter declaration, constant definitions and decision variable 
declarations can be interleaved, but for readability we suggest to put them in the 
order given above. Comments are preceded by `\$' and run to the end of the line.

Parameter values are defined in a separate file, the 
{\em parameter file}. Parameter files have the same header 
as problem models and hold a list of parameter definitions.
Table \ref{tab:modelstructure} gives an overview of the model
structure of problem and parameter files.
Each model part will be discussed in more detail in the following sections.

\begin{table}
\begin{center}
\begin{tabular}{|l||l|}
\hline
Problem Model Structure & Parameter File Structure \\
\hline
\hline
{\tt language ESSENCE' 1.0}  & {\tt language ESSENCE' 1.0}  \\
\ & \\
\$ {\em parameter declaration}       & \$ {\em parameter instantiation} \\
{\tt given n : int}         & {\tt letting n=7} \\
\$ {\em constant definition}        & \\
{\tt letting c=5 }  & \\
\ & \\
\$ {\em variable declaration }       & \\
{\tt find x,y : int(1..n) } & \\
\ & \\
\$ {\em constraints}                 & \\
{\tt such that } & \\
\ \ {\tt x + y >= c,}    & \\  
\ \ {\tt x + c*y = 0}    & \\  
\hline
\end{tabular}
\caption{Model Structure of problem files and parameter files in \eprime. `\$' denotes comments.}\label{tab:modelstructure}
\end{center}
\end{table}

\subsection{Parameter Declarations with {\tt given}}

Parameters are declared with the {\tt given} keyword followed 
by a domain the parameter ranges over. Parameters are allowed to 
range over the infinite domain {\tt int}, or domains that contain an open range
such as {\tt int(1..)} and {\tt int(..10)}. For example,

\begin{verbatim}
given n : int(0..)

given d : int(0..n)
\end{verbatim}

declares two parameters, and the domain of the second depends on the value of
the first. Parameters may have integer, boolean or matrix domains. 

Now we have parameters we can generalise the Sudoku model to represent the 
problem class of all Sudoku puzzles. The parameter is the \texttt{clues} 
matrix, where blank spaces are represented as \texttt{0} and non-zero entries
in \texttt{clues} are copied to \texttt{M}. 

\begin{verbatim}
language ESSENCE' 1.0

given clues : matrix indexed by [int(1..9), int(1..9)] of int(0..9)

find M : matrix indexed by [int(1..9), int(1..9)] of int(1..9)

such that

forAll row : int(1..9) .
    forAll col : int(1..9) .
        (clues[row, col]!=0) -> (M[row, col]=clues[row, col]),

forAll row : int(1..9) .
    allDiff(M[row,..]),

forAll col : int(1..9) .
    allDiff(M[..,col]),	 

$ all 3x3 subsquare have to be all-different
$ i,j indicate the top-left corner of the subsquare. 
forAll i,j : int(1,4,7) .
    allDiff([ M[k,l]  | k : int(i..i+2), l : int(j..j+2)])
\end{verbatim}

\subsection{Constant Definitions with {\tt letting}\label{sec:letting}}

In most problem models there are re-occurring constant values and
it can be useful to define them as constants. The {\tt letting}
statement allows to assign a name with a constant value. The statement

\begin{verbatim}
letting NAME = A
\end{verbatim}

introduces a new identifier {\tt NAME} that is associated with 
the expression \texttt{A}. Every subsequent occurrence of 
{\tt NAME} in the model is replaced by the value of {\tt A}. Please note 
that {\tt NAME} cannot be used in the model {\em before} it has been 
defined. \texttt{A} may be any integer, boolean or matrix expression that does
not contain decision variables. Some integer examples are shown below. 

\begin{verbatim}
given n : int(0..)
letting c = 10
letting d = c*n*2
\end{verbatim}

Here the second integer constant depends on the first. As well as integer and
boolean expressions, lettings may contain matrix expressions, as in the
example below. When using a matrix literal the domain is optional -- the two 
lettings below are equivalent.  The version with the matrix domain may be useful when
the matrix is not indexed from 1. 

\begin{verbatim}
letting cmatrix = [ [2,8,5,1], [3,7,9,4] ]
letting cmatrix2 : matrix indexed by [ int(1..2), int(1..4) ] of int(1..10) 
    = [ [2,8,5,1], [3,7,9,4] ]
\end{verbatim}

Finally new matrices may be constructed using a slice or comprehension, as in the example below
where the letting is used for the table of a table constraint. 

\begin{verbatim}
letting xor_table = [ [a,b,c] | a : bool, b : bool, c : bool, 
                                (a /\ b) \/ (!a /\ !b) <-> c ]
find x,y,z : bool
such that
table([x,y,z], xor_table)
\end{verbatim}

\subsubsection{Constant Domains}

Constant domains are defined in a similar way using the keywords {\tt be domain}.

\begin{verbatim}
letting c = 10
given n : int(1..)
letting INDEX be domain int(1..c*n)
\end{verbatim}

In this example {\tt INDEX} is defined to be an integer domain, the upper bound of
which depends on a parameter \texttt{n} and another constant \texttt{c}. 

Constant domains are convenient when a domain is reused several times. In the 
Sudoku running example, we could use a letting for the domain \texttt{int(1..9)}:

\begin{verbatim}
language ESSENCE' 1.0
letting range be domain int(1..9)

given clues : matrix indexed by [range, range] of int(0..9)

find M : matrix indexed by [range, range] of range

such that

forAll row : range .
    forAll col : range .
        (clues[row, col]!=0) -> (M[row, col]=clues[row, col]),

forAll row : range .
    allDiff(M[row,..]),

forAll col : range .
    allDiff(M[..,col]),	 

$ all 3x3 subsquare have to be all-different
$ i,j indicate the top-left corner of the subsquare. 
forAll i,j : int(1,4,7) .
    allDiff([ M[k,l]  | k : int(i..i+2), l : int(j..j+2)])
\end{verbatim}

\subsection{Decision Variable Declaration with {\tt find}}

Decision variables are declared using the {\tt find} keyword followed by a name and their
corresponding domain. The domain must be finite. The example below
\begin{verbatim}
find x : int(1..10)
\end{verbatim}
defines a single decision variable with the given domain. 
It is possible to define several variables in one \texttt{find} by giving multiple
names, as follows. 

\begin{verbatim}
find x,y,z : int(1..10)
\end{verbatim}

Matrices of decision variables are declared using a matrix domain, as in the following example. 

\begin{verbatim}
find m : matrix indexed by [ int(1..10) ] of bool
\end{verbatim}

This declares \texttt{m} as a 1-dimensional matrix of 10 boolean variables. Simple and matrix
domains are described in Section~\ref{sec:types-domains}.

In the Sudoku running example, we have been using the following two-dimensional matrix domain.

\begin{verbatim}
find M : matrix indexed by [int(1..9), int(1..9)] of int(1..9)
\end{verbatim}

\subsection{Constraints on Parameters with {\tt where}}

In some cases it is useful to restrict the values of the parameters. This is
achieved with the \texttt{where} keyword, which is followed by a boolean expression
containing no decision variables. In the following example, we require the first
parameter to be less than the second. 

\begin{verbatim}
given x : int(1..)
given y : int(1..)
where x<y
\end{verbatim}

\subsection{Objective}
The objective of a problem is either {\tt maximising} or 
{\tt minimising} an integer or boolean expression. For instance,

\begin{verbatim}
minimising x
\end{verbatim}

states that the value assigned to variable $x$ will be minimised.
Only one objective is allowed, and it is placed after all \texttt{given}, 
{\tt find} and {\tt letting} statements.

\subsection{Solver Control}

In addition to instructing the solver to minimise or maximise some expression, 
\eprime also supports some rudimentary options for controlling which variables 
the solver will branch on, and which variable ordering heuristic it will use. 
The information is passed on only when Minion is used as the solver.
These statements are experimental and may be removed from the language in future
versions. 

The \texttt{branching on} statement specifies a sequence of variables for the 
solver to branch on. The \texttt{heuristic} statement specifies the heuristic 
used on only the variables in the \texttt{branching on} list. \texttt{heuristic} 
is followed by \texttt{static}, \texttt{sdf}, \texttt{conflict} or \texttt{srf} and
these options are passed through to Minion. 

The example below tells the solver to branch on \texttt{w} and \texttt{x} using the smallest domain first heuristic.
It will subsequently branch on all the other decision variables using the default 
(static) ordering.

\begin{verbatim}
find w,x,y,z : int(1..10)
branching on [w,x]
heuristic sdf
\end{verbatim}

The \texttt{branching on} statement is followed by a comma-separated list of
individual decision variables or matrices. This list may contain matrices of 
different dimensions and sizes. Decision variables in matrices are enumerated in 
the order produced by the \texttt{flatten} function. 

The \texttt{branching on} list may contain the same decision variable
more than once. This can be useful to pick some variables from a matrix to
branch on first, then include the rest of the matrix simply by including the entire matrix.
In the following example the diagonal of \texttt{M} is branched first, then the rest
of \texttt{M} is included. 

\begin{verbatim}
find M : matrix indexed by [int(1..9), int(1..9)] of int(1..9)

branching on [ [ M[i,i] | i : int(1..9)], M ]
\end{verbatim}

When a matrix is included in the \texttt{branching on} list, it is converted to
a one-dimensional matrix using \texttt{flatten}. 

Optimisation is only performed on the variables in the \texttt{branching on} list. 
Using {\tt branching on} can cause {\tt maximising} and {\tt minimising} to 
function in an unusual way: they will only maximise or minimise on the variables in 
the {\tt branching on} list, and therefore may not return the overall maximal/minimal 
solution.

%

\subsection{Constraints}

After defining constants and declaring decision variables and parameters,
constraints are specified with the keywords {\tt such that}. The constraints in
\eprime are boolean expressions as described in Section~\ref{sec:int-bool-expressions}.

Typically the constraints are written as a list of boolean expressions separated
by the \texttt{,} operator.  


\section{Undefinedness in \eprime}\label{sec:undef}

Since the current version of \eprime is a closed language, there are a finite set
of partial functions in the language. For example, \texttt{x/y} is a partial function
because it is not defined when \texttt{y=0}. In its current version \eprime implements the 
relational semantics as defined by Frisch and Stuckey 
(The Proper Treatment of Undefinedness in Constraint Languages, in Proc.\ Principles and Practice of Constraint Programming - CP 2009, pages 367-382).
The relational semantics has the advantage that it can be implemented efficiently. 

The relational semantics may be summarised as follows:
\begin{itemize}
\item Any integer or matrix expression directly containing an undefined expression is itself undefined.
\item Any domain or domain expression directly containing an undefined expression is itself undefined.
\item Any statement in the preamble (\texttt{find}, \texttt{letting} etc) directly containing an undefined expression is undefined. 
\item Any boolean expression that directly contains an undefined expression is \texttt{false}. 
\end{itemize}

Informally, the relational semantics confines the effect of an undefined expression
to a small part of the problem instance (which becomes \texttt{false}), in many cases avoiding
making the entire problem instance \texttt{false}.

Consider the four examples below. Each contains a division by zero which is an undefined
integer expression. In each case the division is contained in a comparison. Integer
comparisons are boolean expressions. 

\begin{verbatim}
(x/0 = y) = false
(x/0 != y) = false
! (x/0 = y) = true
! (x/0 != y) = true
\end{verbatim}

Applying the rules of the relational semantics results in each of the comparisons inside the brackets 
becoming \texttt{false}:

\begin{verbatim}
(false) = false
(false) = false
! (false) = true
! (false) = true
\end{verbatim}

In the relational semantics, \texttt{(x/0 != y)} is not semantically equivalent to \texttt{!(x/0 = y)},
which is somewhat counter-intuitive. 

Another counter-intuitive case arises with matrix indexing. In the following example,
the expression \texttt{M[0]} is undefined because 0 is not in the index domain. 
If the matrix is boolean (i.e.\ \texttt{DOM} is \texttt{bool}) then \texttt{M[0]}
becomes \texttt{false}, and the model has a solution when \texttt{M[1]=false}.
However, if the matrix contains integer variables (i.e.\ \texttt{DOM} is \texttt{int(0..1)})
then the constraint \texttt{M[0] = M[1]} becomes \texttt{false} and the model has no solutions. 

\begin{verbatim}
find M : matrix indexed by [int(1)] of DOM
such that
M[0] = M[1]
\end{verbatim}

In the \savilerow implementation of \eprime, all partial functions are removed
in a two-step process, before any other transformations are applied. The first step
is as follows. For each partial function a boolean 
expression is created that is \texttt{true} when the partial function is
defined and \texttt{false} when it is undefined. There are six operators that may be
partial: division, modulo, power, factorial, matrix indexing and matrix slicing. 
Table~\ref{tab:undef} shows the generated boolean expression for each operator. 
The boolean expression is then added to the model by connecting it (with \verb1/\1) onto the closest boolean
expression above the partial function in the abstract syntax tree.

\begin{table}  
    \begin{center}
    \begin{tabular}{l|l|l} 
Partial Function & Defined When \\
\hline
{\tt X/Y} & \texttt{Y!=0} \\
\hline
{\tt X\%Y} & \texttt{Y!=0} \\
\hline
{\tt X**Y} & \verb1(X!=0 \/ Y!=0) /\ Y>=01 \\
\hline
\texttt{factorial(X)} & \texttt{X>=0} \\
\hline
Matrix indexing:   &  $\mathtt{I}\in\mathtt{D1}$ \verb1/\1  \\
\texttt{M[I,J,K]}  &  $\mathtt{J}\in\mathtt{D2}$ \verb1/\1  \\
where domain of \texttt{M} is  &  $\mathtt{K}\in\mathtt{D3}$  \\
\texttt{matrix indexed by [D1, D2, D3] of DBase}   &  \\
\hline
Matrix slicing:   &  $\mathtt{I}\in\mathtt{D1}$ \verb1/\1  \\
\texttt{M[I,..,K]}  &   $\mathtt{K}\in\mathtt{D3}$  \\
where domain of \texttt{M} is &    \\
\texttt{matrix indexed by [D1, D2, D3] of DBase}  &  \\
\end{tabular}
\end{center}
    \caption{Partial functions in \eprime. \texttt{X}, \texttt{Y}, \texttt{I}, \texttt{J} and \texttt{K} are arbitrary expressions
    of the correct type.}
\label{tab:undef}
\end{table}

The second step is to replace the partial function \texttt{P} with a total function \texttt{SP}.
For each input where \texttt{P} is defined, \texttt{SP} is defined to the same value. For inputs where \texttt{P} is 
undefined, \texttt{SP} takes a default value (typically \texttt{0} for integer expressions).

Once both steps have been applied to each partial function, the model is well
defined everywhere. This is done first, before any other model transformations,
and thus allows all subsequent transformations to be simpler because there is no
need to allow for undefinedness. 
For example, the expression \texttt{x/y=x/y} may not be simplified to \texttt{true},
because it is \texttt{false} when \texttt{y=0}. However, after replacing the partial
division function, the resulting equality can be simplified to \texttt{true}.  In
general any equality between two syntactically identical expressions can be simplified
to \texttt{true} once there are no partial functions.


\appendix

\section{Operator Precedence in \eprime}\label{app:op}

Table \ref{tab:precedence} shows the precedence and associativity of operators
in \eprime. As you would expect, operators with higher precedence are applied first.

Left-associative operators are evaluated left-first, for example {\tt 2/3/4 = (2/3)/4}.
The only operator with right associativity is {\tt **}. This allows double
exponentiation to have its conventional meaning: {\tt 2**3**4 = 2**(3**4)}

Unary operators usually have a higher precedence than binary ones. There is one
exception to this rule: that \texttt{**} has a higher precedence than unary minus. 
This allows \texttt{-2**2**3} to have its conventional meaning 
of \texttt{-(2**(2**3))=-256}, as opposed to \texttt{(-2)**(2**3)=256}.

\begin{table}
\begin{center}
\begin{tabular}{l|lcc}
Operator & Functionality & Associativity & Precedence \\
\hline
{\tt !} & Boolean negation &  & 20\\
{\tt ||} & Absolute value  &  & 20\\
\hline
{\tt **} & Power & Right & 18 \\
\hline
{\tt -} & Unary negation &   & 15\\
\hline
{\tt *} & Multiplication & Left & 10 \\
{\tt /} & Division & Left & 10 \\
{\tt \%} & Modulo & Left & 10 \\
\hline
{\tt intersect} & Domain intersection & Left & 2 \\
\hline
{\tt union} & Domain union & Left & 1 \\
{\tt +} & Addition & Left & 1 \\
{\tt -} & Subtraction \& Domain Subtraction & Left & 1 \\
\hline
{\tt =} & Equality & Left & 0 \\
{\tt !=} & Disequality & Left & 0 \\
{\tt <=} & Less-equal & Left & 0 \\
{\tt <} & Less than & Left & 0 \\
{\tt >=} & Greater-equal & Left & 0 \\
{\tt >} & Greater than & Left & 0 \\
{\tt <=lex} & Lex less-equal & Left & 0 \\
{\tt <lex} & Lex less than & Left & 0 \\
{\tt >=lex} & Lex greater-equal & Left & 0 \\
{\tt >lex} & Lex greater than & Left & 0 \\
\hline
{\tt in} & Value in a set & Left & 0 \\
\hline
{\tt \verb1/\1} & And & Left & -1 \\
\hline
{\tt \verb1\/1} & Or & Left & -2 \\
\hline
{\tt ->} & Implication & Left & -4 \\
{\tt <->} & If and only if & Left & -4 \\
\hline
{\tt forAll, exists, sum} & Quantifiers & & -10 \\
\hline
{\tt ,} & And & Left & -20 \\
      \end{tabular}
\end{center}
    \caption{Operator precedence in \eprime}

\label{tab:precedence}
\end{table}

\section{Reserved Words} \label{app:reservedwords}

The following words are keywords and therefore are not allowed to be used as identifiers. 

\begin{verbatim}
forall, forAll, exists, sum,  
such, that, letting, given, where, find, language, 
int, bool, union, intersect, in, false, true
\end{verbatim}

\bibliographystyle{plain}
\bibliography{general}

\end{document}